\documentclass[sigconf,screen,leftjustify]{acmart}

\copyrightyear{2024}
\acmYear{2024}
\setcopyright{rightsretained}
\acmConference[KDD '24]{Proceedings of the 30th ACM SIGKDD Conference on Knowledge Discovery and Data Mining}{August 25--29, 2024}{Barcelona, Spain} \acmBooktitle{Proceedings of the 30th ACM SIGKDD Conference on Knowledge Discovery and Data Mining (KDD '24), August 25--29, 2024, Barcelona, Spain}\acmDOI{10.1145/3637528.3671463}
\acmISBN{979-8-4007-0490-1/24/08}



\usepackage{tabularray} 
\usepackage{xcolor}
\usepackage[section]{placeins}
\usepackage{tikz}
\tikzstyle{edge} = [draw, thick, -latex]
\usepackage{rotating}

\FloatBarrier
\tikzset{
  basic/.style  = {draw, rounded corners=6pt, text width=3cm, drop shadow, font=\sffamily, rectangle},
  root/.style   = {basic, rounded corners=6pt, thin, align=center,
                   fill=cyan!30},
  level 2 red/.style = {basic, rounded corners=6pt, thin,align=center, fill=blue!25 , 
                   text width=5em},
level 2 green/.style = {basic, rounded corners=6pt, thin,align=center, fill=blue!25,
                   text width=16em},
  level 3 red/.style = {basic,rounded corners=6pt, thin, align=center, fill=blue!5,
                 text width=7em},
level 3 white/.style = {     draw=none,
drop shadow={opacity=0},thin,align=center, fill=white,text width=6em,       },
}
\definecolor{light-gray}{HTML}{E5E4E2}

\usetikzlibrary{decorations.pathreplacing,arrows,shapes,positioning,shadows,trees}

\begin{document}

\title{Decoding the AI Pen: Techniques and Challenges in Detecting AI-Generated Text}
\author{Sara Abdali}

\email{saraabdali@microsoft.com}
\affiliation{%
  \institution{Microsoft}
  \city{Redmond}
  \state{WA}
  \country{USA}
}
\author {Richard Anarfi}
\authornote{Equal contribution\label{note1}}
\email{ranarfi@microsoft.com}
\affiliation{%
  \institution{Microsoft}
  \city{Cambridge}
  \state{MA}
  \country{USA}
}

\author {CJ Barberan\footref{note1}}
\email{cjbarberan@microsoft.com}
\affiliation{%
  \institution{Microsoft}
  \city{Cambridge}
  \state{MA}
  \country{USA}
}
\author{Jia He\footref{note1}}
\email{hejia@microsoft.com}
\affiliation{%
  \institution{Microsoft}
  \city{Cambridge}
  \state{MA}
  \country{USA}
}


\begin{abstract}
Large Language Models (LLMs) have revolutionized the field of Natural Language Generation (NLG) by demonstrating an impressive ability to generate human-like text. However, their widespread usage introduces challenges that necessitate thoughtful examination, ethical scrutiny, and responsible practices. In this study, we delve into these challenges, explore existing strategies for mitigating them, with a particular emphasis on identifying AI-generated text as the ultimate solution. Additionally, we assess the feasibility of detection from a theoretical perspective and propose novel research directions to address the current limitations in this domain.
\end{abstract}


\maketitle
\section{Introduction}
Large Language Models (LLMs) constitute a transformative advancement in the field of natural language processing (NLP). Their applications traverse a wide spectrum of domains, including question answering~\cite{Zaib2021BERTCoQACBC,Bhat2023InvestigatingAO}, sentiment analysis~\cite{Batra2021BERTBasedSA,Kheiri2023SentimentGPTEG} and specially text generation~\cite{Senadeera2022ControlledTG}. As LLMs are trained on extensive textual corpora, they exhibit a remarkable capacity to produce human-like text.
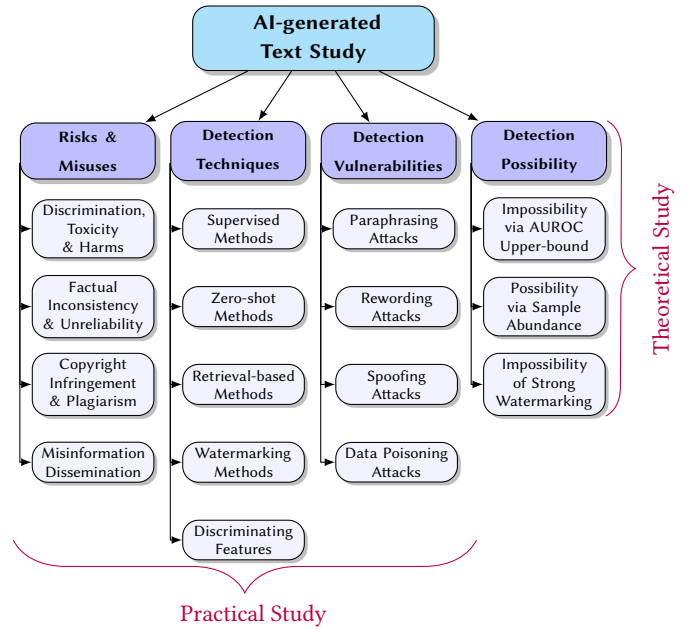
\begin{figure}[t!]
\begin{tikzpicture}[
   level distance=1.5cm,
  level 1/.style={sibling distance=20mm},
  edge from parent/.style={->,draw},
  >=latex]

\node[root] {\textbf{ \small AI-generated Text Study}}
  child {node[level 2 red] (c1) {\scriptsize \textbf{Risks \& Misuses}}}
  child {node[level 2 red ] (c2) {\scriptsize \textbf{Detection Techniques}}}
  child {node[level 2 red] (c3) {\scriptsize \textbf{Detection Vulnerabilities}}}
  child {node[level 2 red] (c4) {\scriptsize \textbf{Detection Possibility}}};

\begin{scope}[every node/.style={level 3 red}]

\scriptsize
\node [below of = c1, xshift=2pt, yshift=-1pt] (c11) {Discrimination, Toxicity \& Harms};
\node [below of = c11,yshift=-1pt] (c12) {Factual Inconsistency \& Unreliability};

\node [below of =c12,
yshift=-1pt] (c13) {Copyright Infringement \& Plagiarism};

\node [below of = c13,yshift=-1pt] (c14) {Misinformation Dissemination};

\node [below of = c2,xshift=2pt, yshift=-1pt] (c21) {Supervised Methods};

\node [below of = c21 ,yshift=-1pt](c22) {Zero-shot Methods};

\node [below of = c22, yshift=-1pt] (c23) {Retrieval-based Methods};

\node [below of = c23,yshift=-1pt]
(c24) {Watermarking Methods};
\node [below of = c24,yshift=-1pt]
(c25) {Discriminating Features};

\node [below of = c3,xshift=2pt, yshift=-1pt] (c31) {Paraphrasing Attacks};

\node [below of = c31,xshift=1pt, yshift=-1pt] (c32) {Rewording Attacks};

\node [below of = c32,yshift=-1pt]
(c33) {Spoofing Attacks};

\node [below of = c33,yshift=-1pt]
(c34) {Data Poisoning Attacks};

\node [below of = c4,xshift=2pt, yshift=-1pt]
(c41) {Impossibility via AUROC Upper-bound};
\node [below of = c41, yshift=-1pt]
(c42) {Possibility via Sample Abundance};
\node [below of = c42, yshift=-1pt]
(c43) {Impossibility of Strong Watermarking};

\end{scope}

\begin{scope}[every node/.style={level 3 white}]
\node [below of = c14, yshift=-1pt]
(c15) {};

\node [below of = c34, yshift=-1pt]
(c35) {};
\end{scope}

\foreach \value in {1,2,3,4}
  \draw[->] (c1.190) |- (c1\value.west);

\foreach \value in {1,2,3,4,5}
  \draw[->] (c2.190) |- (c2\value.west);

\foreach \value in {1,2,3,4}
  \draw[->] (c3.190) |- (c3\value.west);

\foreach \value in {1,2,3}
  \draw[->] (c4.190) |- (c4\value.west);

\draw[decorate, decoration={brace, amplitude=11pt},color=purple]
(c4.north east) -- (c43.south east) node [midway, xshift=0.5cm, right]{\begin{sideways}Theoretical Study\end{sideways}} ;

\draw[decorate, decoration={brace, amplitude=20pt, mirror},color=purple]
(c15.west) -- (c35.east) node [midway, yshift=-1cm,  xshift=1.2cm, left] {Practical Study};
\end{tikzpicture}

\caption{\centering An overview of responsible AI-generated text study, with an emphasize on detection approaches and their challenges.}
\label{fig:crownjewel}
\end{figure}
 
\par However, the ubiquity of LLMs brings forth a convergence of opportunities and challenges that necessitate prudent examination. Among these challenges, we encounter the potential for LLMs to produce biased, toxic, or harmful content~\cite{kuchnik2023validating,Deshpande2023ToxicityIC,Weidinger2021EthicalAS,Wen2023UnveilingTI}. Additionally, there are concerns related to intellectual property rights infringement~\cite{peng2023are,stokel2022ai} and the misuse of LLMs for malicious purposes, such as disseminating misleading information and propaganda~\cite{Vykopal2023DisinformationCO,Mozes2023UseOL}. These multifaceted considerations highlight the need for judicious evaluation and ethical management of LLMs across different contexts.
\par Leveraging AI-generated text could be regarded as an efficacious strategy for alleviating the aforementioned challenges. However, the convergence of AI-generated content with human-written texts has reached a point where discerning between the two has become increasingly intricate. The task of distinguishing LLM-generated text from human-written content presents a dual challenge. 
On one hand, identifying disparities can enhance the quality of AI-generated material. On the other hand, this endeavor also complicates the identification process. In recent years, researchers have proposed various methodologies for detecting AI-generated text to contribute to the ongoing exploration of this landscape.
\par This work aims to comprehensively investigate responsible AI-generated text. To this end, first, we highlight the risks and misuses associated with such text, while also discussing common mitigation strategies. As a significant solution, we thoroughly explore AI-generated text, concentrating on thematic categorization and assessing their constraints and vulnerabilities. Additionally, we conduct a theoretical exploration to assess the feasibility and potential of detecting AI-generated text. By adopting a theoretical lens, we seek to determine whether such detection is achievable or if detection remains an elusive goal within the generative AI domain.
\par The rest of this paper is organized as follows: In section~\ref{sec:risks}, we discuss the potential risks arising from the misuse of LLMs. Section~\ref{sec:text-detect} provides a comprehensive categorization of various text detection strategies, emphasizing their role as mitigation techniques. Subsequently, in Section~\ref{sec:vulns}, we describe the vulnerabilities inherent in these strategies. Moving forward, Section~\ref{sec:possibility} engages in a theoretical exploration of the feasibility of detection, followed by section~\ref{sec:newops} where we propose new avenues of research. Finally, in section~\ref{sec:conclusions}, we conclude. Figure \ref{fig:crownjewel} demonstrates an overview of the paper.
\section{Risks and Misuse of AI-Generated Text}
  \label{sec:risks}
LLMs harbor the capacity to generate harmful content and enable malicious actions. These include spreading toxic, biased, or harmful language, misinformation propagation and even committing plagiarism.
In the subsequent sections, we will explore a non-comprehensive list of potential risks associated with LLM misuse, while discussing proposed mitigation techniques.

 \subsection{Discrimination, Toxicity, and Harms}
LLMs, have the potential to produce text that exhibits discriminatory, offensive, or harmful characteristics towards individuals or groups. The manifestation of such undesirable language depends on factors such as the quality and diversity of the training data used, the design decisions made during model development, and the intended or unintended contexts in which the model is applied~\cite{gehman2020realtoxicityprompts,Deshpande2023ToxicityIC,Cui2023FFTTH}. Thus, LLMs may pose ethical and social challenges that require careful evaluation and regulation.
\par A recent work published by DeepMind~\cite{Weidinger2021EthicalAS}, structures the risk landscape associated with LLMs. It outlines six specific risk areas, including discrimination, exclusion and toxicity, and discusses the potential mitigation approaches and challenges. For discrimination, exclusion and toxicity, the paper suggests improving data quality and diversity, applying fairness metrics and interventions, and implementing content moderation and reporting mechanisms.

\par In another study by Deshpande et al.,~\cite{Deshpande2023ToxicityIC} it is shown that ChatGPT, when assigned a persona, can exhibit significant toxicity. Particularly, this risk is elevated for vulnerable groups such as students, minors, and patients. This work emphasizes that toxicity is closely tied to the style of communication, with explicit negative prompts leading to increased toxicity.

\par Furthermore, the research reveal that specific genders and ethnicities are disproportionately susceptible to encountering toxic content. 
Deshpande et al. posit that this phenomenon arises from an over-reliance on reinforcement learning with human feedback (RLHF) as a mechanism to mitigate model toxicity. However, the feedback provided to the model may carry inherent biases. For instance, feedback related to toxicity concerning different genders might be influenced by the representation of those genders.
\par Kour et al.~\cite{Kour2023UnveilingSV} recently introduced the AttaQ dataset, designed to provoke harmful or inappropriate responses from LLMs. They conduct evaluations on multiple LLMs using this dataset and observe that, in numerous instances, LLMs generate unsafe outputs.

\par More importantly, LLMs have the potential to produce implicit toxic responses that elude existing classifiers. These harmful outputs, while not easily identifiable, can offend individuals or groups by insinuating negative or false statements. As a result, this undermines the safety and reliability of NLG systems and gives rise to ethical concerns. 
\par A recent study by Wen et al.~\cite{Wen2023UnveilingTI} investigates how LLMs generate implicit toxic content that conventional toxicity classifiers struggle to detect. The proposed approach leverages reinforcement learning to uncover and expose this implicit toxicity, emphasizing the need to fine-tune classifiers using annotated examples derived from the attack method for enhanced detection capabilities.

\par Given the multifaceted nature of content generation by LLMs, it is critical to explore the intricate interplay of user-driven, data-driven, and model-driven factors that contribute to the production of toxic and harmful content. Therefore, further research is warranted to comprehensively investigate the deleterious effects and toxicity associated with LLMs. Future research endeavor should focus on devising robust methodologies and mechanisms for prevention, detection, and mitigation. By doing so, we not only bolster the safety and reliability of LLMs but also make significant contributions to the advancement of related disciplines.
\vspace{-4pt}
\subsection{Factual Inconsistency and Unreliability of AI Responses}
Enforcing factual coherence during reasoning constitutes a key challenge for LLMs. These models often demonstrate tendencies such as overlooking conditions, misinterpreting context, and even hallucinating content in response to specific queries~\cite{Laban2023LLMsAF}.

\par For example, an investigation into GPT-3 by Khatun et al.~\cite{Khatun2023ReliabilityCA} revealed that while the model adeptly avoids blatant conspiracies and stereotypes, it hesitates when handling commonplace misunderstandings and debates. In particular, the model’s responses exhibit variability across different questions and contextual scenarios, emphasizing the inherent unpredictability of GPT-3.

\par Zhou et al.~\cite{Zhou2024RelyingOT} recently conducted a study revealing that LLMs, including ChatGPT and Claude, exhibit deficiencies in conveying uncertainties when responding to questions. Furthermore, these models occasionally display unwarranted overconfidence, even when their answers are incorrect. While LLMs can be coerced to express confidence, this process is fraught with high error rates.
\par Significantly, the study highlights that users encounter challenges in assessing the accuracy of LLM-generated responses. Their judgment is influenced by the tone and style of the LLMs, which may introduce bias. This issue is extremely important, as biases against uncertain text may impact the LLM training and evaluation. 

\par To mitigate such mistakes, various strategies have been proposed through fine-tuning~\cite{lewkowycz2022solving,rajani-etal-2019-explain,zelikman2022star}, prompt engineering techniques like verification, scratchpads ~\cite{Cobbe2021TrainingVT,nye2022show}, Chain-of-Thought (CoT)~\cite{wei2022chain}, Reversing Chain-of-Thought (RCoT) ~\cite{Xue2023RCOTDA}, RLHF~\cite{Ziegler2019FineTuningLM,NIPS2017_d5e2c0ad}, iterative self-reflection~\cite{shinn2023reflexion,madaan2023selfrefine}, self-consistency~\cite{wang2023selfconsistency}, society of minds strategy~\cite{Du2023ImprovingFactuality}, pruning truthful datasets~\cite{christiano2023deep}, adjusting the system parameters to limit model creativity~\cite{Muneeswaran2023MinimizingFI},
external knowledge retrieval~\cite{Guu2023RALM} and training-free methods with likelihood estimation~\cite{kadavath2022language}.
\par The existing body of research emphasize the unspeakable capabilities of LLMs, while also recognizing their proneness to errors. Thus, it is essential to approach LLM outputs with caution.
\subsection{Copyright Infringement and Plagiarism}

LLMs could pose a substantial risk to academic writing by elevating the likelihood of copyright violations and plagiarism. For instance, writers might employ LLMs to produce articles without original composition, while students could resort to LLMs for completing homework assignments. These practices erode academic integrity and subvert the objectives of assignments and examinations~\cite{khalil2023chatgpt,stokel2022ai}, 

\par In response to this challenge, researchers have devised several detectors aimed at discerning between text authored by humans and that generated by AI. These detectors fall into two main categories: black-box methods, exemplified by studies such as~\cite{Wang2023M4MM,Quidwai2023BeyondBB,Liu2023CheckMI}, and white-box detection, as explored by~\cite{Vasilatos2023HowkGPTIT}. 

\par In the black-box detection approach, there is limited access to the LLM’s output text. Specifically, we interact with the LLM via its API, which provides us with the generated text. Black-box detectors rely on collecting samples of both human-written and AI-generated text. These samples are then used to train a classifier.
The trained classifier discriminates between LLM-generated and human-written texts based on features extracted from the text samples~\cite{Quidwai2023BeyondBB,Liu2023CheckMI,GPTZero}.

\par In contrast to the black-box approach, the white-box approach necessitates additional access to the probabilities associated with each token in the model~\cite{Vasilatos2023HowkGPTIT}. Consequently, there are fewer white-box detectors currently available. Thus, black-box methods that are independent of model access and can be readily adjusted to a new model seem to be more feasible and practical.

\par Another crucial aspect to consider is the generalizability of AI-generated text detectors. These detectors should perform effectively on unseen data across diverse dimensions, including: models, languages and domains. Achieving robustness and reliability across these dimensions contributes to the overall effectiveness of AI-generated text detectors in real-world scenarios.
\par  Their study involved using various generative models to create text articles and then distinguishing between AI-generated and human-written content. They applied traditional machine learning techniques and transformer-based models to analyze stylistic features. Interestingly, while these methods performed well within their specific domains, they struggled with out-of-domain detection tasks. Furthermore, their findings indicate that when text detectors are trained on content generated by one LLM and then tested on data produced by a different LLM, performance tends to decline and generalizability becomes an issue. However, given that this study examines only a limited number of detection methods, more comprehensive and systematic assessments are necessary to validate this aspect of LLM capabilities. In section~\ref{sec:text-detect}, we will delve further into the topic of AI-generated text detection techniques.

\subsection{Misinformation Dissemination}

LLMs, especially when integrated into Open-Domain Question Answering (ODQA) systems, can inadvertently contribute to the creation and dissemination of misinformation~\cite{MisinformationPan,Chen2023CanLL,pan-etal-2023-risk}. An intuitive approach, as suggested by Pan et al.~\cite{pan-etal-2023-risk} to counteract the spread of misinformation in ODQA systems is to reduce its prevalence. In fact, the goal is to minimize the proportion of misinformation that these QA systems encounter. Achieving this involves retrieving more context paragraphs to provide a solid background for readers.
\par However, research have shown that expanding the context size has minimal impact on mitigating the performance degradation caused by misinformation . Consequently, the straightforward strategy of diluting misinformation by increasing context size proves ineffective for defending against it~\cite{pan-etal-2023-risk}. 
\par A more straightforward approach involves prompting LLMs to issue warnings regarding misleading information. For instance, readers could be advised: “Be cautious, as some of the text may be intentionally misleading” .

\par Furthermore, it is feasible to detect and filter out misinformation produced by LLMs using different characteristics, such as content, style, or propagation structure. Chen et al.~\cite{Chen2023CanLL}, for example, introduce four instruction-tuned strategies to enhance LLMs for misinformation detection. One of these strategies is \textit{Instruction Filtering}, which aims to exclude LLM outputs that deviate from instructions or contain misleading content, \textit{Instruction Verification} which verifying the outputs of the LLM against the instructions or external sources to check their validity and reliability and \textit{Instruction Combination} which combines multiple instructions to generate more diverse and accurate outputs from the LLM.
\par Another interesting approach introduced by Chen et al.~\cite{Chen2023CanLL} is the \textit{Reader Ensemble} technique. This method leverages multiple LLMs or other models to cross-check the accuracy and coherence of information produced by a specific LLM. Chen and colleagues also suggest employing \textit{Vigilant Prompting}, which provides precise prompts or instructions to LLMs. These instructions serve a dual purpose: preventing the generation of misinformation and disclosing the model’s machine identity.
\par As AI-generated texts increasingly blend seamlessly with human-written content, the demand for more effective methods to detect misleading information produced by AI grows. In the next section, we will dive deeper into detection methods.
\section{AI-generated Text Detection Techniques}
\label{sec:text-detect}
 In the previous section, we briefly explored a classification of detection techniques into two primary categories: black-box and white-box techniques. As mentioned, in the black-box scenario, the access is limited to the output text produced by LLM given an arbitrary input. Conversely, in the white-box context, we gain additional access into the model’s output probabilities for individual tokens. In this section, we discuss various detection techniques, considering both their strengths and vulnerabilities.
\subsection{Supervised Detection} A frequently employed detection strategy is fine-tuning a language model on datasets comprising both AI-generated and human-written texts~\cite{Solaiman2019ReleaseSA,Bakhtin2019RealOF,Antoun2023TowardsAR,Zhan2023G3DetectorGG,Li2023DeepfakeTD}. The majority of LLMs require substantial computational resources, rendering it exceedingly challenging to curate sufficiently large datasets that comprise a diverse spectrum of samples. As a result, this approach is not generally an optimal solution. Moreover, it is susceptible to adversarial attacks, including data poisoning~\cite{Chen2017TargetedBA,Schwarzschild2020JustHT,Yang2021BeCA}. 
\par For example, malicious actors have the potential to elude detection by leveraging their access to human.written texts present in the training set and detector rankings. Moreover, within a white-box context, attackers can undermine detector training—a concerning scenario, especially considering that numerous detectors rely on commonly used datasets, making them susceptible even to basic attacks. Additionally, these techniques are prone to the paraphrasing attack, where a paraphrased layer is added to the generative text model, allowing deception of any detector, including those utilizing supervised neural networks~\cite{Krishna2023ParaphrasingED,Sadasivan2023CanAT}.

\subsection{Zero-shot Detection} Another line of research employ pre-trained models as zero-shot classifiers to discern text written by AI, eliminating the necessity for supplementary training or data collection~\cite{Su2023DetectLLMLL,ZeroGPT,Wang2023BotOH,Gehrmann2019GLTRSD}. According to~\cite{Mitchell2023DetectGPTZM}, commonly employed techniques often set a threshold for the predicted per-token log probability to identify LLM-generated texts. This approach is grounded in the observation that passages generated by AI often exhibit a negative log probability curvature.
\par While this method mitigates the risk of data poisoning attacks and minimizes data and resource requirements, it remains vulnerable to other attacks such as spoofing~\cite{Shayegani2023SurveyOV} and paraphrasing~\cite{Krishna2023ParaphrasingED,Sadasivan2023CanAT}.

\subsection{Retrieval-based Detection}An alternative avenue of research utilize information retrieval methods to differentiate between texts written by humans and those generated by machines. This is achieved by comparing a given text with a database of texts created by LLMs and identifying semantically similar matches~\cite{Krishna2023ParaphrasingED,Sadasivan2023CanAT}. 
\par Nevertheless, these approaches are impractical for real-world use due to their reliance on an extensive and up-to-date database of AI-generated texts. Such databases can be computationally costly or may not be available across all domains, tasks, or models. Additionally, like other detection techniques we discussed earlier, these methods are susceptible to paraphrasing and spoofing attacks~\cite{Krishna2023ParaphrasingED,Sadasivan2023CanAT,Wolff2020AttackingNT,LIANG2023100779}.
\subsection{Watermarking-based Detection}
\label{sec:watermarking}
Another alternative avenue, referred to as \textit{watermarking} techniques, employs a model signature within the generated text outputs to imprint specific patterns. 
\par For instance, Kirchenbauer et al.~\cite{kirchenbauer2023watermark} propose a soft watermarking approach that categorizes tokens into green and red lists, facilitating the construction of these patterns. In this approach, a watermarked LLM selects a token, with a high likelihood, from the green list based on its preceding token. Remarkably, these watermarks often remain imperceptible to human observers.
\par To better understand the technique proposed by Kirchenbauer et al., assume an autoregressive language model is trained on a vocabulary $V$ of size $|V|$. Given a sequence of tokens as input at step $t$, a language model predicts the next token in the sequence by outputting a
vector of logit scores $l_t \in R^
{|V|}$ with one entry for each item in the vocabulary. A random number
generator is seeded with a context window of $h$ preceding tokens, based on a pseudo-random function (PRF) $f : N
h \rightarrow N$. With this random seed, a subset of tokens of size $\gamma|V|$, where $\gamma \in (0,1)$  is green list size,
are “colored green” and
denoted $G_t$. Now, the logit scores $l_t$ are modified such that with a hardness parameter $\sigma >0$:
\begin{equation*}
l_{tk}= 
\begin{cases}
    l_{tk}+\sigma,& \text{if } k\in G_t\\
    l_{tk},              & \text{otherwise}
\end{cases}
\end{equation*}
In a straightforward scenario the process involves passing scores through a softmax layer and then sampling from the resulting output distribution. This tends to introduce a bias toward tokens from the set $G_t$. Once watermarked text is generated, it is possible to verify the watermark even without direct access to the LLM. This verification is achieved by recomputing the $G_t$ at each position and identifying the set of token positions associated with the $G_t$. For $T$ tokens, the statistical significance of the watermark is assessed by z-score:
\begin{equation*}
    z=\frac{(|S|-\gamma T)}{\sqrt{\gamma(1-\gamma)T}}
\end{equation*}
When this z-score is large (and the corresponding P-value is small), one can be confident that the text
is watermarked~\cite{kirchenbauer2023watermark}.

\par Despite previous indications, watermark-based techniques remain both theoretically and practically vulnerable to rewording attacks. Notably, even LLMs safeguarded by watermarking schemes are susceptible to spoofing attacks. In these attacks, human adversaries inject their own text into human-generated content, creating an illusion that the material originated from LLMs~\cite{Sadasivan2023CanAT}. The interested reader is referred to~\cite{Sadasivan2023CanAT} for more details.

\par Moreover, unless all highly successful LLMs are uniformly safeguarded, watermarking remains an ineffective strategy for preventing LLM exploitation. \par Furthermore, the practical applicability of watermarking is restricted, especially in scenarios where only black-box language models are accessible. Due to API providers withholding probability distributions for commercial reasons, most third-party developers creating API-based applications lack the capability to independently watermark text.

\par To tackle this challenge, Yang et al.~\cite{Yang2023WatermarkingTG} have developed a watermarking framework to empower third parties with the ability to autonomously inject watermarks into black-box language model scenarios. In this approach, a binary encoding function is leveraged that generates random binary encodings corresponding to individual words. These encodings adhere to a Bernoulli distribution, where the probability of a word representing bit-1 is approximately $0.5$. To embed a watermark, this distribution  is modified by selectively replacing words associated with bit-0 using contextually relevant synonyms that represent bit-1. Subsequently, a statistical test is employed to detect the watermark.

\par Another study conducted by Kirchenbauer et al.~\cite{Kirchenbauer2023OnTR} explores the effectiveness of watermarks in identifying AI-generated text. They assess how watermarked text withstands challenges such as human restructuring, non-watermarked LLM paraphrasing, and seamless integration into longer handwritten documents.  
\par They observe that even after undergoing automated and human paraphrasing, watermarks can still be discerned. When a sufficient number of tokens are identified, paraphrased versions tend to inadvertently leak n-grams or even larger segments of the original text. Hence, high-confidence detection remains possible despite these attacks weakening the effectiveness of the watermark. The study proposes interpreting watermarking reliability as a function of text length. Surprisingly, even when attempting to remove the watermark intentionally, human writers struggle to do so if the text exceeds 1,000 words. This interpretation emerges as a significant characteristic of watermarking. According to this work, watermarking remains the most dependable strategy, as alternative paradigms like retrieval and loss-based detections have not demonstrated substantial improvements with increasing text length.

\par Despite prior findings, a recent study by Zhang et al.~\cite{Zhang2023WatermarksIT} demonstrates that, under some assumptions, no robust watermarking scheme can prevent an attacker from removing the watermark without significantly degrading the output quality. We will further discuss this finding in section~\ref{sec:Watermarking_impos}.

\subsection{Detection via Discriminating Features}
Another stream of work is to identify and classify text based on distinguishing features. For instance, Yu et al.~\cite{Yu2023GPTPT} have discovered a genetic inheritance characteristic in GPT-generated text. Leveraging this characteristic, the model’s output becomes a rearrangement of content from its training corpus. In other words, when repeatedly answering questions, the model’s responses are constrained by the information within its training data, resulting in limited variations. This hypothesis suggests that the output of an LLM, such as ChatGPT, is predictable. Therefore, for highly similar questions, the model tends to produce correspondingly similar responses.
\par Drawing an analogy, paternity testing utilizes DNA profiles to determine whether an individual is the biological parent of another person. This process becomes particularly crucial when legal rights and responsibilities related to parenthood are in question, and uncertainty exists regarding the child’s paternity.
\par In another study, Yang et al.~\cite{Yang2023DNAGPTDN} introduce a training-free detection strategy known as Divergent N-Gram Analysis (DNA-GPT). This approach assesses the dissimilarities between a given text and its remaining truncated portions using n-gram analysis in black-box scenarios or probability divergence in white-box scenarios.
 \par For the black box scenario, Yang et al. define DNA-GPT BScore:
\begin{multline*}
BScore(S,\Omega) =\\
\frac{1}{K}\Sigma_{k=1}^K \Sigma_{n=n_0}^N f(n) \frac{| \text{\scriptsize n-grams}(\hat{S}_k)\cap \text{\scriptsize n-grams}(S_2)|}{|\hat{S}_k|| \text{\scriptsize n-grams}(S_2)|}
\end{multline*}
where $S$ is the LLM output, $S_2$ the human written ground truth, $f(n)$ a weight function for different n-grams and $\Omega=\{\hat{S}_1,\dots, \hat{S}_K\}$

\par For white-box scenario, they propose calculating a DNA-GPT WScore between  $\Omega$ and $S$:
\begin{equation*}
    WScore(S,\Omega)=\frac{1}{K}\Sigma_{k=1}^K log\frac{ p(S_2|S_1)}{p(\hat{S}_k|S_1)}
\end{equation*}
Where $\Omega$ is a set of $K$ samples of an LM decoder and $\hat{S}=LM(S_1)$ and $S_2$ is the human written ground truth.
“In both black-box and white-box scenarios, two parameters significantly impact detection accuracy: the truncation ratio and the number of re-prompting iterations $K$. This strategy reveals substantial discrepancies between AI-generated and human-written texts. It highlights substantial disparities between AI-generated and human-written texts.
\par Another distinguishing feature is the vulnerability of text to manipulations. Both AI-generated and human-written texts can be adversely affected by minor alterations, such as word replacements. However, recent research have revealed that AI-generated text is particularly prone to such manipulations~\cite{Mitchell2023DetectGPTZM,Su2023DetectLLMLL}.
\par For instance, Su et al.~\cite{Su2023DetectLLMLL} propose a metric called the Log-Likelihood Log-Rank Ratio (LRR) to quantify the sensitivity of LLMs to perturbations:
\begin{equation*}
    LPR= -\frac{\Sigma_{i=1}^{t} \log p_\theta(x_i|x_{<i})}{\Sigma_{i=1}^{t} \log r_\theta(x_i|x_{<i})} 
\end{equation*}
where  $r_\theta(x_i|x_{<i}) \geq 1$  is the rank of token $x_i$ conditioned on the previous tokens~\cite{Su2023DetectLLMLL}.
The Log-Likelihood in the numerator reflects the absolute confidence for the correct token, whereas the Log-Rank in the denominator considers relative confidence. Together, they provide complementary information about the texts.
\par They also propose Normalized Log-Rank Perturbation (NPR):
\begin{equation*}
    \text{NPR}=\frac{\frac{1}{n} \Sigma_{p=1} ^n \log r_\theta(\tilde{x}_p)}{\log r_\theta(x)}
\end{equation*}
where small perturbations are applied on the target text $x$ to produce the perturbed text $\tilde{x}_p$. 
\par This work demonstrates that the LRR tends to be larger for AI-generated text, making it a useful discriminator between AI and human-generated content. One plausible explanation is that, in AI-generated text, the log rank stands out more prominently than the log likelihood, resulting in a distinct pattern that LRR captures.
\par The rationale behind NPR lies in the fact that both AI-generated and human-written texts experience adverse effects from minor alterations. Specifically, the log rank score tends to increase after perturbations. However, AI-generated text is particularly vulnerable to such alterations, resulting in a more pronounced increase in the log rank score following perturbation. As a result, this pattern suggests a higher NPR score for AI-generated texts~\cite{Su2023DetectLLMLL}.
\par Table~\ref{tab:AI-text-detection} illustrates an overview of detection strategies, highlighting the
 vulnerabilities associated with each category.
\section{Vulnerabilities of Detection Techniques}
\label{sec:vulns}
Despite various methods mentioned above, most categories of detection strategies are susceptible to paraphrasing or spoofing attacks. To tackle this, retrieval-based detectors have been employed as detection strategies but these detectors store the output of LLMs in a database and perform semantic searches to extract the best matches. Although it enhances the detector’s resilience against paraphrasing attacks, there are privacy concerns related to storing user-LLM conversations. Additionally, this technique is ineffective when dealing with recursive paraphrasing~\cite{Sadasivan2023CanAT}. 
\par Moreover, researchers have discovered that by effectively optimizing prompts, LLMs can successfully evade many detection techniques. For example, in a recent work~\cite{Lu2023LargeLM}, Lu et al. propose a novel method called Substitution-based In-Context example Optimization (SICO). In SICO, discriminating features are extracted from both human and AI-generated texts. These features, along with a paraphrasing prompt, are combined and fed to the LLM to modify its output. The prompt is carefully optimized through word and sentence replacements, aiming to minimize the chances of detection while maximizing the similarity between human and AI-generated texts. Evaluation results strongly demonstrate the vulnerability of existing methods. 

\par
\begin{figure}[!t]
\includegraphics[width=1\linewidth]{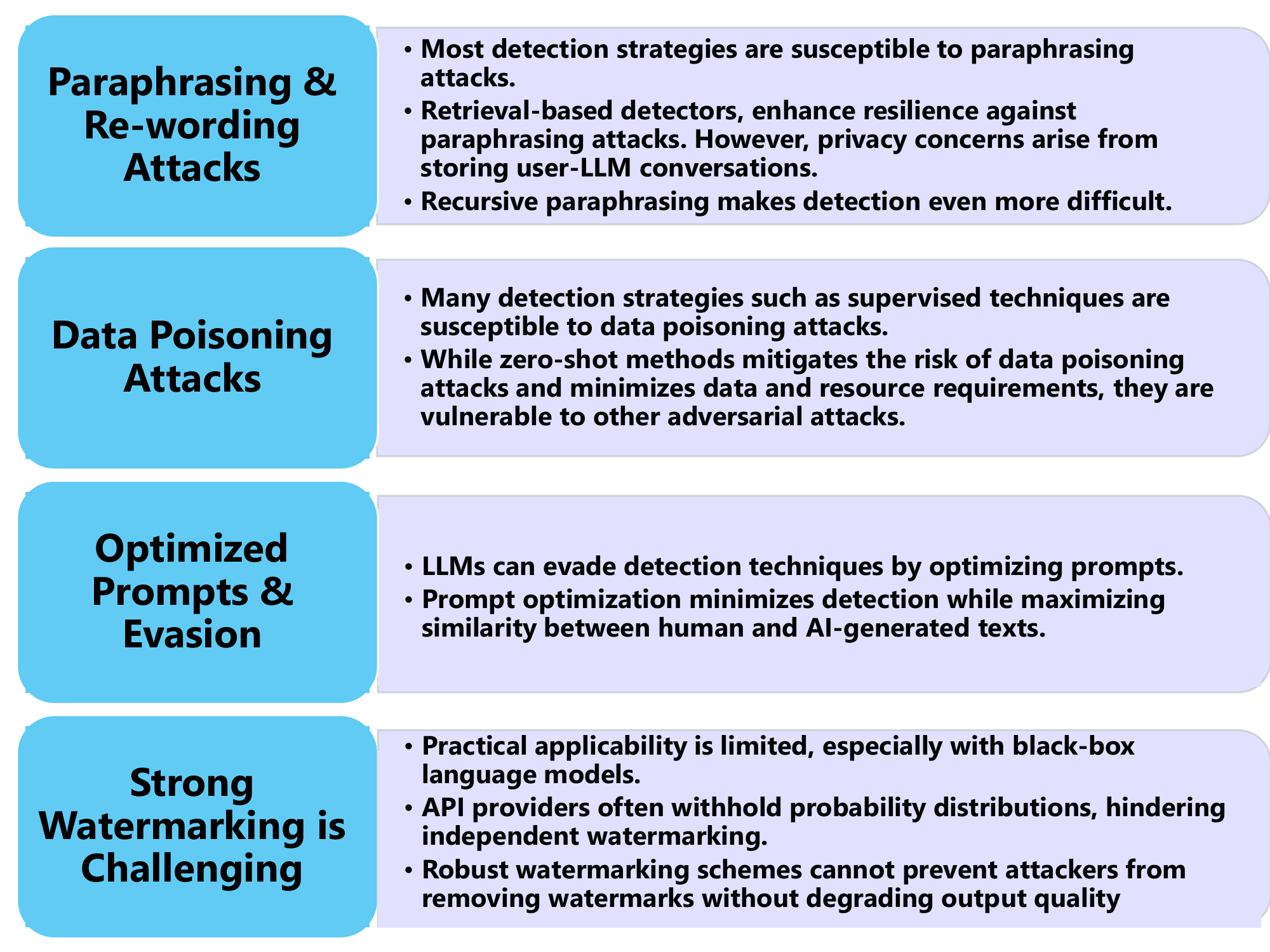}
\caption{A summary of detection vulnerabilities.}
\label{fig:vuls}
\end{figure}

 Despite being considered an effective detection strategy, watermarking faces several challenges. Firstly, unless all LLMs are uniformly safeguarded, watermarking remains ineffective. Secondly, its practical applicability is limited, especially when dealing with black-box language models. Thirdly, API providers often withhold probability distributions, preventing third-party developers from independently watermarking text. Lastly, recent research suggest that no robust watermarking scheme can prevent attackers from removing watermarks without significantly degrading output quality. Therefore, watermarking generative models might be fundamentally unachievable, and alternative approaches are necessary to protect the intellectual property of model developers and LLM users. Figure~\ref{fig:vuls} presents a concise overview of vulnerabilities in detection techniques.
\section{Detection Possibility Through a Theoretical Lens}
\label{sec:possibility}
\par Given the increasing interest in detecting text generated by LLMs, researchers have recently investigated the theoretical aspects of this task. They explore the fundamental feasibility and limitations associated with identifying LLM-generated text. In this section we explore some of the notable research studies on this topic.
\label{fig:vuln}
\subsection{Impossibility via AUROC Upper-bound}
\par
\begin{figure}[t]
\includegraphics[width=1\linewidth]{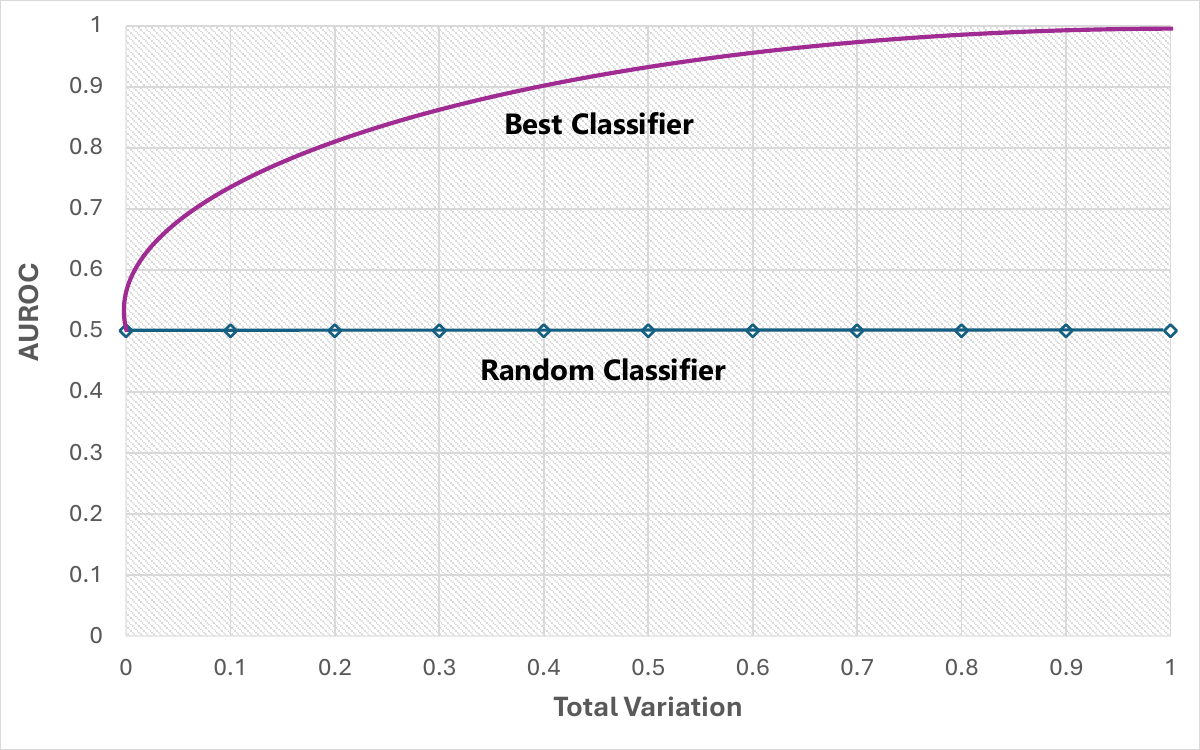}
\caption{Comparing AUROC
of the optimal detector to a random classifier demonstrates that as the TV distance between AI and human text distributions reduces,
the AUROC of the optimal detector also decreases accordingly.}
\label{fig:auroc}
\end{figure}

 In a new study, Sadasivan et al. 
 present a discovery of impossibility~\cite{Sadasivan2023CanAT}. They claim that:
 \newline
  \newline
 \noindent\fcolorbox{blue!60}{blue!5}
{

\minipage[t]{\dimexpr1\linewidth-4\fboxsep-3\fboxrule\relax} As language models become increasingly sophisticated and adept at emulating human text, the effectiveness of even the best-possible detectors diminishes significantly. 
\endminipage
}
\newline
\newline
Sadasivan et al. establish an upper-bound for the area under the Receiver Operating Characteristic (ROC) curve of any decoder $\mathcal{D}$:
\begin{equation*}
    AUROC(\mathcal{D})\leq\frac{1}{2}+TV(\mathcal{M},\mathcal{H})-\frac{TV(\mathcal{M},\mathcal{H})^2}{2}
    \label{eq:upperbound}
\end{equation*}
where $TV(\mathcal{M},\mathcal{H})$ is the total variation distance between machine and human-written texts distributions. This formula indicates that
when human and machine generated texts are very similar i.e., $TV(\mathcal{M},\mathcal{H})$ is very small, even the most effective detector may exhibit only marginal improvement over a random classifier. Figure~\ref{fig:auroc} illustrates how the above bound grows as a function of the total variation distance. 

\subsection{Possibility via Sample Abundance}
\par Another interesting study by Chakraborty et al.~\cite{Chakraborty2023OnTP} suggests that:
\newline
  \newline
 \noindent\fcolorbox{blue!60}{blue!5}
{
\minipage[t]{\dimexpr1\linewidth-4\fboxsep-3\fboxrule\relax} 
As long as the distributions of human-generated and AI-generated texts are not identical (which is typically the case), it remains feasible to detect AI-generated texts. 
This detection becomes possible when we gather sufficient samples from each distribution. 
\endminipage
}
\newline
\newline
Interestingly, Chakraborty et al. demonstrate that the AUROC curve, as proposed by Sadasivan et al., might be overly conservative for practical detection scenarios. Specifically, they introduce a hidden possibility by replacing $TV(\mathcal{M} ,\mathcal{H})$ with $TV(\mathcal{M}^{\bigotimes n},\mathcal{H}^{\bigotimes n})$ in AUROC equation, where $m^ {\bigotimes n} := m \bigotimes m \bigotimes \dots \bigotimes m$ (n times) denotes the product distribution over sample set $\mathcal{S}:=\{s_i\}, i\in \{1,\dots n\}$, as does $h^{\bigotimes n}$. Thus, for any detector $\mathcal{D}$, with a given collection of i.i.d. samples either from human 
or AI, the AUROC curve is defined as:
\begin{equation*}
    AUROC(\mathcal{D})\leq\frac{1}{2}+TV(\mathcal{M}^{\bigotimes n},\mathcal{H}^{\bigotimes n})-\frac{TV(\mathcal{M}^{\bigotimes n},\mathcal{H}^{\bigotimes n})^2}{2}
    \label{eq:upperbound}
\end{equation*}

where $TV(\mathcal{M}^{\bigotimes n},\mathcal{H}^{\bigotimes n})= 1-exp(-nl_c(m,h)+o(n)$ and $l_c(m,h)$ is the Chernoff information.
\par Since $TV(\mathcal{M}^{\bigotimes n},\mathcal{H}^{\bigotimes n})$ is an increasing sequence, it eventually converges to $1$ as the number of samples for each distribution increases. 
Thus, the upper bound of AUROC increases exponentially with respect to the number of samples.
It is clear that if this happens, the total variation distance approaches $1$ quickly, and hence increasing the AUROC. Interested readers are encouraged to refer to~\cite{Chakraborty2023OnTP} for further details.

\subsection{Impossibility of Strong Watermarking}
\label{sec:Watermarking_impos}
\par In another work, Zhang et al.~\cite{Zhang2023WatermarksIT}, investigate the theoretical aspect of watermarking detection. They conceptualize watermarking as the process of embedding a statistical signal a.k.a. ``watermark'' into a model’s output. This embedded watermark serves as a signal for later verification, ensuring that the output is indeed originated from the model. A robust watermarking prevents an attacker from erasing the watermark without causing significant quality degradation.
\par Zhang et al. put forth two fundamental assumptions:
    \paragraph{``Quality Oracle'':} grants the attacker access to an oracle capable of evaluating the quality of outputs. This oracle assists the attacker in assessing the quality of modified responses.
    \paragraph{``Perturbation Oracle'':}  allows the attacker to modify an output while maintaining a nontrivial probability of preserving quality.
\vspace{4pt}
\par  The perturbation oracle essentially induces an efficiently mixing random walk on high-quality outputs. Their investigation culminates in a compelling result: 
 \newline
 \newline
\noindent\fcolorbox{blue!60}{blue!5}
{

\minipage[t]{\dimexpr1\linewidth-4\fboxsep-3\fboxrule\relax} 
 ``\textbf{\Large Given} a prompt $p$ and a watermarked output $y$, for every public or secret-key watermarking setting that satisfying these assumptions, there \textbf{\Large exists} an efficient attacker that can leverage the quality and perturbation oracles to obtain an output ${y}^{'}$ with a probability very close to $1$. The attacker’s goal is to: \\ \textbf{\Large find} an output ${y}^{'}$ s.t. \\ (1) ${y}^{'}$ is not
 watermarked with high probability and, \\(2) ${Q(p,{y}^{'} )} \geq {Q(p,y)}$''~\cite{Zhang2023WatermarksIT}.

\endminipage
}
\newline
\newline
  In simpler terms, watermarking without causing significant quality degradation is impossible. Thus, Zhang et al. propose alternative approaches to safeguard the intellectual property of developers.

\par Detecting AI-generated text is an essential and complex task and current state-of-the-art methods occasionally face limitations due to a lack of comprehensive understanding regarding the fundamental feasibility and boundaries of this task. Thus, it is crucial to continue exploring and investigating the theoretical aspects of the matter.
 \section{Limitations and Future Research}
 \label{sec:newops}
This paper offers a comprehensive overview of the latest advancements and recommended practices in detecting AI-generated text. However, existing methods often face limitations and are susceptible to malicious attacks. Thus, it is essential to continue exploring both theoretical and practical aspects of this realm. In light of this, we propose opportunities for advancing the field of responsible AI, vulnerability assessment of AI-generated text detection studies. Some of these opportunities include:
\paragraph{\bf Curating diverse and representative datasets} is crucial for training and evaluating detection models. The existing datasets may not comprise all the potential types and sources of AI text. Specially, it is essential to curate a dataset containing outputs from various generative models.
    \paragraph {\bf Further investigating interpretable features} can help discern the subtle differences between human-written and AI-generated text, especially given the limited availability of such features. Additionally, investigating their vulnerability to adversarial attacks is crucial.
    \paragraph{\bf Exploring advanced and adaptable learning techniques} to effectively address the dynamic and ever-evolving nature of AI-generated text. These methods include adversarial learning, meta-learning, and self-supervised learning~\cite{WeberWulff2023TestingOD}.
    \paragraph{\bf A comprehensive multi-aspect evaluation of detection techniques against adversarial attacks} as literature lacks such study to investigate relative vulnerabilities of detection methods. Such an analysis should consider various perspectives, including efficacy across different models and resilience against adversarial attacks. This study can aid in identifying techniques suitable for specific scenarios.
     \paragraph{\bf Developing hybrid detection strategies} that combine features and techniques to enhance robustness and adaptability. For example, one can create a hybrid approach by integrating watermarking and feature-based techniques.
   \paragraph{\bf A comprehensive understanding of the fundamental feasibility and boundaries} as there is often an absence of a thorough grasp of the fundamental feasibility and limitations within current state-of-the-art methods. Therefore, there is a need for deeper exploration and investigation into the theoretical aspects of this task to facilitate the creation of more resilient and efficient techniques while also uncovering novel avenues for research. 

\section{Conclusions}
\label{sec:conclusions}
In this paper, we delve into a thorough study of AI-generated text. Our analysis comprises not only the potential risks and misuses associated with content generated by LLMs, but also investigates widely recognized techniques for mitigating these risks. As a pivotal mitigation strategy, we conduct a comprehensive study of AI-generated text detection techniques, categorizing them into five distinct categories and meticulously comparing their weaknesses and vulnerabilities. Our investigation approaches AI-generated text detection from both empirical and theoretical angles, shedding light on the intricacies of the field, which leads to proposing new avenues of research to bolster this critical area of study.
\section{Acknowledgements}
This study represents independent research conducted by the authors and does not necessarily represent the views or opinions of any organizations. We express our gratitude to the anonymous reviewers for their valuable and constructive feedback.
\begin{table*}
\centering
\small
\caption{AI-generated text detection techniques and their vulnerabilities.} 
\label{tab:AI-text-detection}
\begin{tblr}[!t]
{
colspec = {@{}p{3cm}p{2.25cm}p{5.5cm}p{5.5cm}},
row{1} = {bg=blue!20},
row{2} = {bg=blue!3},
row{3} = {bg=blue!3},
row{4} = {bg=blue!3},
row{5} = {bg=blue!3},
row{6} = {bg=blue!3},}
\hline
\textbf{\centering Related Papers}&\textbf{{Method}}&\textbf{\centering Main Idea}&\textbf{\centering Vulnerabilities} \\
\hline
\cite{Solaiman2019ReleaseSA,Bakhtin2019RealOF,Antoun2023TowardsAR,Zhan2023G3DetectorGG,Li2023DeepfakeTD}& Supervised \newline detection&To fine-tune a model on sets of AI and
human generated texts.& Training on commonly
used datasets, makes it vulnerable to most attacks including paraphrasing.\\
\hline
~\cite{Su2023DetectLLMLL,ZeroGPT,Mitchell2023DetectGPTZM,Gehrmann2019GLTRSD,Wang2023BotOH,Guo2023AuthentiGPTDM}& Zero-shot\newline detection&To use a pre-trained language model in zero-shot settings.& Reduces the risk of data poisoning attacks and eliminates data and resource over-
heads, but it is still susceptible to other adversarial
attacks like spoofing and paraphrasing.\\
\hline
~\cite{Krishna2023ParaphrasingED,Sadasivan2023CanAT,Wolff2020AttackingNT,LIANG2023100779}& Retrieval-based \newline detection& To apply methods of information retrieval to match a given text with a collection of texts generated by LLMs and finding similarities in meaning.& It is impractical because it requires a large and updated collection of texts, which is computationally expensive, or may be unavailable for all domains, tasks or models. It is also vulnerable to paraphrasing and spoofing attacks.\\
\hline
\cite{kirchenbauer2023watermark,Yang2023WatermarkingTG,Kirchenbauer2023OnTR,Sadasivan2023CanAT}& Watermarking&To use a model signature in the produced
text outputs to stamp particular pattern.& The most trustworthy strategy, but is shown to be fundamentally impossible for generative models.  It is susceptible to attacks such as rewording and spoofing.\\
\hline
\cite{Yu2023GPTPT,Yang2023DNAGPTDN,Mitchell2023DetectGPTZM,Su2023DetectLLMLL,Zhang2023WatermarksIT} & Feature-based \newline detection&To identify and classify based on extracted discriminating features.&Susceptible to adversarial attacks such as paraphrasing. \\
\hline
\end{tblr}
\end{table*}

\end{document}